\documentclass[runningheads]{llncs}

\usepackage[T1]{fontenc}
\usepackage{graphicx}
\usepackage{comment}
\usepackage{amsmath}
\usepackage{amssymb}

\begin{document}
\title{Perspective-Aware AI in Extended Reality}
%
%


\author{
Daniel Platnick\inst{1,2} \and
Matti Gruener\inst{1} \and
Marjan Alirezaie\inst{1,2} \and \\
Kent Larson\inst{3} \and
Dava J. Newman\inst{3} \and
Hossein Rahnama\inst{1,2,3}
}
\institute{
Flybits Labs, Creative Ai Hub, Canada
\and
Toronto Metropolitan University, Canada
\and
MIT Media Lab, USA\\
\email{\{daniel.platnick,matti.gruener,marjan.alirezaie\}@flybits.com}\\
\email{\{kll,dnewman,rahnama\}@mit.edu}
}





%
\authorrunning{Daniel Platnick et al.}
%
%

\maketitle
\renewcommand\thefootnote{} 
\footnote{This version of the article has been accepted for publication, after peer review, and is subject to Springer Nature’s AM terms of use, but is not the Version of Record and does not reflect post-acceptance improvements, or any corrections. The Version of Record will be available at: http://dx.doi.org/[insert DOI].
}
\renewcommand\thefootnote{\arabic{footnote}}

\begin{abstract}
AI-enhanced Extended Reality (XR) aims to deliver adaptive, immersive experiences—yet current systems fall short due to shallow user modeling and limited cognitive context.
We introduce Perspective-Aware AI in Extended Reality (PAiR), a foundational framework for integrating Perspective-Aware AI (PAi) with XR to enable interpretable, context-aware experiences grounded in user identity.
PAi is built on \emph{Chronicles}—reasoning-ready identity models learned from multimodal digital footprints that capture users’ cognitive and experiential evolution.
PAiR employs these models in a closed-loop system linking dynamic user states with immersive environments.
We present PAiR’s architecture, detailing its modules and system flow, and demonstrate its utility through two proof-of-concept scenarios implemented in the Unity-based OpenDome engine.
PAiR opens a new direction for human-AI interaction by embedding perspective-based identity models into immersive systems.
\end{abstract}

\section{Introduction} \label{sec:intro}
Perspective-Aware AI (PAi) is an approach in human-AI interaction that enables individuals to digitally experience the world through perspectives other than their own \cite{PAi1-2024,PAi-2021}.
PAi systems have the potential to greatly improve critical decision-making, personalized simulations, and human-AI collaboration in many domains such as financial advising, education, digital twin simulations, and Extended Reality (XR) \cite{PAi2-2024,Sai2024twins,Radianti2020VR-Education}.
At its core, PAi is a computational framework designed to capture and reason over human perspectives in a structured, interpretable, and context-aware manner  \cite{PAi3-2024}.
Unlike conventional personalization methods that rely on user analytics and predictive recommendations, PAi systems represent user identity with a data structure referred to as a \emph{Chronicle}.
A \emph{Chronicle} is a reasoning-ready user identity model that represents the cognitive, behavioral, and experiential evolution of an individual or social entity by learning a dynamic knowledge graph from their available digital footprint data. 
This footprint comprises the user's unique interaction history, decision patterns, and situational contexts across multiple modalities over time \cite{PAi1-2024,PAi3-2024}.

Compared to conventional AI approaches, PAi systems generate more contextualized user experiences by querying personalized \emph{Chronicles} to understand user perspectives, beliefs, and identities.
PAi systems also support the sharing of \emph{Chronicles} to synthesize experiences grounded in the perspectives of others who have permissibly contributed their data to the system.
By modeling user perspectives, PAi surpasses reactive personalization and enables peer-to-peer systems that are more inclusive, less biased, and grounded in diverse viewpoints \cite{PAi2-2024}.

One of the most promising applications of PAi is in XR, which includes Virtual Reality (VR), Augmented Reality (AR), and Mixed Reality (MR) \cite{AlhakamyXR-encompass}. 
AI-driven XR applications today leverage reinforcement learning for personalization, user modeling for adaptive experiences, digital twins for real-world simulation, and generative AI for AI-assisted world-building \cite{Pasupuleti2025XRVRAR,Pat1999usermod,Ratican2023-genai,Schwarz2025meta}. 
Despite these advances in AI-powered XR systems, critical limitations persist:

1) Current personalization methods in XR are primarily reactive, optimizing short-term engagement rather than adapting meaningfully to individual cognitive preferences and their evolution over time \cite{10.3389/frvir.2022.1015155}.

2) User modeling approaches typically rely on surface-level behavioral cues and physiological states, neglecting deeper cognitive dynamics related to user identity, and behaviors \cite{10.1007/2023}.

3) Digital twins commonly concentrate on physical fidelity and operational prediction, overlooking cognitive and social dimensions crucial for comprehensive simulations \cite{GAFFINET2025104230}.

4) Generative AI methods often produce static narratives and interactions disconnected from dynamic user perspectives \cite{57dca900c67c4572bc5c185827d53c63}.


To address these challenges, we introduce Perspective-Aware AI in Extended Reality (PAiR), a framework unifying PAi with XR to enable perspective-aware immersive experiences. Our key contributions include:

\begin{enumerate}
    \item We propose PAiR, a unified framework bridging Perspective-Aware AI (PAi) and Extended Reality (XR) to enable dynamic, deeply personalized immersive interactions grounded in user perspectives. 
    PAiR creates a closed-loop connection between PAi and XR, where immersive experiences are driven by user-specific \emph{Chronicles} that adapt based on evolving user behavior.
    \item We demonstrate PAiR through integration with an XR engine as a proof-of-concept implementation, and present two illustrative scenarios: the Perspe-
    
    ctive-Aware Financial Helper and the Perspective-Aware Desk Environment.
\end{enumerate}

The remainder of this paper is structured as follows: Section \ref{sec:bkgrnd} establishes PAiR within PAi and XR literature. Section \ref{sec:pair} introduces the PAiR framework, outlining its modules and system flow. Section \ref{sec:poc} presents the implementation and scenario-based demonstrations. Section \ref{sec:discussion} concludes with future research directions.

\section{Background}
\label{sec:bkgrnd}

This section reviews relevant literature on Perspective-Aware AI and Extended Reality, highlighting the gaps in current user modeling approaches.

\subsection{Perspective-Aware AI}

Perspective-Aware AI (PAi) is a computational paradigm for modeling identity through dynamic knowledge graphs called \emph{Chronicles} \cite{PAi1-2024}, which capture the evolving behaviors, contexts, and perspectives of individuals or societal entities over time. PAi enables systems to reason over user identity models that reflect each user's unique trajectory across diverse situations. It builds on prior work in user modeling, human-centric AI \cite{PAi-2021,Jiang2024HAI,Amershi2019Guidelines,vanDerMeulen2025ToM}, and context-aware computing, addressing key limitations such as the lack of support for long-term evolution, semantic grounding, and interpretability in identity modeling.


\emph{Chronicles} are constructed from multimodal digital footprints—text, images, and social interactions—collected via decentralized, privacy-preserving channels \cite{PAi2-2024,Kosinski2013DigitalFootprint}. These inputs are embedded using modality-specific encoders and temporally segmented \cite{PAi1-2024}. To semantically ground these representations, PAi uses a predefined ontology called the Situation Graph (SG), encoding high-level concepts like sentiment, emotion, location, and ambiance \cite{PAi3-2024}. The complete construction pipeline is detailed in \cite{PAi1-2024}.


Once built, \emph{Chronicles} can be embedded into downstream systems to support shareable, perspective-aware experiences that adapt to a user's context, history, and personality \cite{PAi3-2024}. 
Leveraging Large Language Models (LLMs) as symbolic translators, PAi maps human or machine language into \emph{Chronicle} queries and transforms \emph{Chronicle} outputs into reasoning-ready language—enabling seamless interaction with structured identity graphs \cite{raffel2023exploringlimitstransferlearning}.
By enabling reasoning over structured identity models, PAi applies broadly to domains requiring adaptive human-AI interaction—such as education, decision support, simulation, and, as explored in this work, immersive Extended Reality.


\subsection{Extended Reality}

Extended Reality (XR) encompasses Virtual, Augmented, and Mixed Reality, and enables interaction in hybrid physical-digital environments \cite{AlhakamyXR-encompass}. 
XR is increasingly used in domains like education, healthcare, and simulation, offering spatial, embodied, and multimodal experiences \cite{Radianti2020VR-Education,Graessler2017DigitalTwinHuman}. 

AI now plays a vital role in shaping XR content, modeling users, and generating interactive narratives, with techniques such as reinforcement learning, emotion detection, and generative models driving adaptive experiences \cite{Marougkas2024VRPersonalization,Pardini2022PersonalizedVR,vanLoon2018VR-Empathy}. However, current AI-powered XR systems struggle to reflect deeper layers of user identity. 
Personalization remains reactive and short-term, focusing on surface-level behaviors like gaze or gesture \cite{Marougkas2024VRPersonalization}. User models often miss internal perspectives, long-term preferences, or evolving beliefs \cite{Tehranchi2023CognitiveUI}, and generative systems rarely ground media content in the user's psychological context.

To advance, XR must enable perspective-aware interaction, where experiences reflect users' values, views, and evolving needs.

\section{PAiR: Perspective-Aware AI in Extended Reality}
\label{sec:pair}

This section introduces PAiR and presents its architectural blueprint for future PAi-enabled XR systems, outlining the core modules and system flow.



\vspace{-0.3cm}

\subsection{PAiR Overview}

PAiR is a foundational framework connecting PAi to XR, allowing immersive systems to reason over user identities represented by \emph{Chronicles}. PAiR comprises two main components: the PAi Component, for reasoning and \emph{Chronicle} interaction, and the XR Component, for spatial context and rendering. Together, they form a closed-loop system where user behavior shapes both the environment and evolving identity.



\begin{figure}[t]
  \centering
    \includegraphics[width=1.0\textwidth]{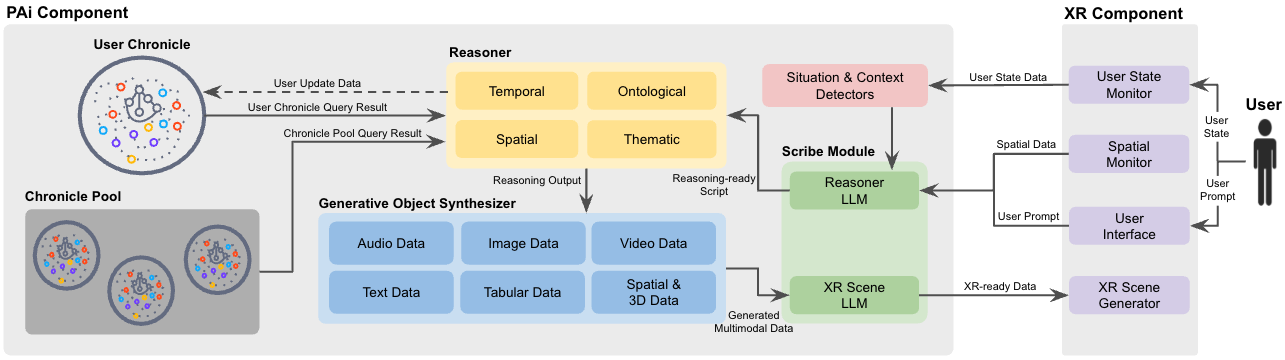}
    \label{fig:pair}
    \caption{An overview of the PAiR architecture. The framework includes a PAi component and an XR component. 
    The figure illustrates PAiR's core modules and data flow, which transform user inputs into personalized XR experiences. 
    The dashed line indicates data flow from user monitoring to continuously update the \emph{Chronicle.}}
\end{figure}

\vspace{-0.3cm}

\subsection{PAiR Modules}
PAiR consists of two primary components: the XR Component and the PAi Component, detailed as follows.

\subsubsection{XR Component.} 
The XR Component manages spatial context, user interactions, and dynamic scene updates to deliver immersive experiences. Designed as a modular, generalizable blueprint, it can be customized to specific applications. PAiR's XR Component includes the following modules:

\begin{itemize}
    \item \textbf{Spatial Monitor:} 
    Provides spatial context by managing the placement, and interaction of virtual and physical entities within the XR environment.
    \item \textbf{User State Monitor:} Monitors and interprets behavioral, physiological, and environmental signals (e.g., heart rate or movement) to provide continuous insight into the user's status and context.
    \item \textbf{User Interface:} An active communication channel allowing users to directly issue commands and queries, or interact with the system through methods like text input, or voice commands.
      \item \textbf{XR Scene Generator:} Dynamically updates the virtual environment by integrating media objects, retrieved or generated from \emph{Chronicles}, with real-time spatial data, ensuring coherent immersive experiences.
\end{itemize}

\subsubsection{PAi Component.} The PAi Component processes user input and contextual cues (e.g., commands, expressions, sentiment) to generate personalized virtual experiences based on \emph{Chronicles}. It includes the following modules:

\begin{itemize}
    \item \textbf{Situation \& Context Detectors:} A collection of off-the-shelf models to interpret sensory data (e.g., facial expressions), producing high-level state classifications such as \emph{happy}, \emph{sad}, or \emph{distracted}.
    \item \textbf{LLM Scribe Module:} 
    A bidirectional translator that converts user requests into machine-readable scripts for the \emph{Reasoner}, and transforms the \emph{Reasoner}’s outputs into scripts for the 3D engine. 
    The Reasoner LLM is fine-tuned to generate structured queries—expressed as triplets or logic-based clauses—that align with the \emph{Chronicle}'s schema. The XR Scene LLM translates symbolic outputs (e.g., semantic graphs, high-level actions, annotated elements) into XR-ready configuration files with spatial layouts, object placements, animations, and environment settings, ensuring the rendered experience reflects both reasoning and user intent.
    \item \textbf{Chronicle:} A reasoning-ready digital representation of an entity that captures its evolving experiences, behaviors, and context, enabling PAiR systems to deliver dynamic, PAi-driven adaptive experiences.
    \item \textbf{Chronicle Pool:} A collection of different shared \emph{Chronicles}, accessible via query by PAiR applications with appropriate consent.
    \item \textbf{Reasoner:} A computational module always associated with \emph{Chronicles} that is aware of their structure and able to query them for a given request. 
    Depending on the request and the \emph{Chronicles}' content, the \emph{Reasoner} can apply various forms of reasoning:
    
    \textbf{Spatial reasoning} involves understanding the physical layout or relative positions of entities within an environment, enabling queries about location, direction, proximity, or movement.
    \textbf{Temporal reasoning} reasons over the timing and sequence of events or states, enabling functionality related to duration, ordering, change over time, and temporal dependencies.
    \textbf{Ontological reasoning} relies on the conceptual hierarchy and relationships between entities, enabling inference based on class membership, inheritance, and type constraints.
    \textbf{Thematic reasoning} identifies patterns and connections based on abstract or semantic themes, enabling understanding of purpose, roles, narratives, or high-level contextual meaning across entities or events.
    
    \item \textbf{Generative Object Synthesizer:} Receives the \emph{Reasoner}'s output and either retrieves existing content (e.g., recalling a memory by displaying a specific picture) or generates a new data object by conditioning on relevant \emph{Chronicle} data (e.g., generating a video of the user climbing a mountain they have never visited, using a photo of them climbing a ladder as conditioning input). 
    This may be accomplished using embeddings or pretrained models specialized for generating the relevant data types, such as images, text, audio, or tabular data. 
\end{itemize}

\vspace{-0.3cm}

\subsection{PAiR System Flow}


In PAiR, interaction begins either directly via the User Interface (e.g., text or voice commands) or passively through the User State Monitoring module, which captures contextual and emotional states via behavioral signals. Monitoring data is processed by state-of-the-art situation and context detectors—such as facial expression and sentiment recognition models—to generate qualitative descriptions of the user’s state (e.g., anxious, joyful, bored). Whether from explicit prompts or inferred context, the input is passed to a fine-tuned LLM that converts it into semantically interpretable scripts for the \emph{Reasoner}.

The \emph{Reasoner}, which also receives spatial data containing anchor point definitions, descriptive metadata, and supported runtime events, uses this initial information to semantically ground symbolic reasoning within spatial references. Additionally, the \emph{Reasoner} applies the descriptions to infer affordances of objects through ontological or thematic reasoning (e.g., recognizing that a desk is suitable for placing items on a wall is appropriate for displaying visuals). Subsequent personalization, context-awareness decisions, and perspective-aware configurations are fully managed by the PAi Component, dynamically populating placeholders based on user-specific contexts and situations. The \emph{Reasoner} then generates queries to extract hyper-personalized information from the \textit{Chronicles}, which contain the behavioral and situational knowledge of each entity.

Based on the \emph{Reasoner}'s output and the content of the \emph{Chronicles}, the \emph{Object Synthesizer} either retrieves existing content or generates new data that is consistent with the user's request and profile. This output is further processed by a fine-tuned LLM that translates the reasoning output into executable scripts for the XR Component. Finally, the XR Scene Generator updates the environment and renders a personalized, immersive experience for the user.

As shown in Figure \ref{fig:pair} (dashed line), user monitoring data can also serve as an additional source of information to update the \emph{Chronicle} based on the user's reactions in different situations. This data is sent to the \emph{Reasoner} to adjust the \emph{Chronicle} content when appropriate thresholds and validated theories are met.

\section{Demonstrating PAiR with OpenDome}
\label{sec:poc}


This section outlines the initial PAiR implementation using OpenDome, a Unity-based XR framework, followed by two use cases: the Perspective-Aware Financial Helper and the Perspective-Aware Desk Environment.




\subsection{PAiR Instantiation in OpenDome}
We implemented PAiR within OpenDome, a Unity-based XR Framework.
OpenDome provides an abstraction layer that separates virtual environments into (a) the structural layout, defined by anchor points, and (b) the dynamic capabilities of these structures, defined by event types and visual elements. 
Each space is equipped with predefined anchor points (e.g., "table center", "wall center") where elements such as 3D charts or memory visualizations can be dynamically placed. These anchor points, along with descriptive metadata, are managed through the \emph{Spatial Monitor} module. 
Each description provides meaningful context, such as "surface for presenting data" or "emotionally neutral background", which can be interpreted by the \emph{Scribe Module} and used by the \emph{Reasoner} to semantically ground symbolic reasoning in the spatial environment.

When OpenDome initializes a runtime session, it triggers PAiR by sending a structured representation of the XR space (i.e., Spatial Data) to the PAi Component. This data includes anchor point definitions, associated descriptive metadata, and supported runtime events, providing the \emph{Reasoner} with the spatial potential and capabilities of the XR environment. 



\subsection{Perspective-Aware Financial Helper}
\label{pafh}

The Perspective-Aware Financial Helper is an immersive environment in OpenDome where users can request and visualize personalized financial advice. 
The following step-by-step example illustrates how PAiR, via OpenDome, transforms user requests and contextual cues into grounded, personalized XR experiences:\\




\subsubsection{Step 1: User Request} \
The user issues a request, converted to a prompt: \texttt{"Show me my credit card spending on the table in front of me."}

\subsubsection{Step 2: Format Request for Reasoning} \
This prompt is then appended with relevant context and spatial data of the XR space. Next, it is input to the Reasoner LLM, which converts the request into a symbolic representation:

{\small
\begin{verbatim}
<situation_1, has_participant, user>
<situation_1, has_intent, visualize_spending_profile>
<situation_1, has_target_entity, credit_card>
<situation_1, has_target_location, table_in_front>
\end{verbatim}
}

\subsubsection{Step 3: Reasoning and Chronicle Querying} \
The \textit{Reasoner} uses its structural understanding of the \emph{Chronicle's} schema to translate situation-describing triples into a semantic query. This translation involves:
\begin{enumerate}
    \item Extract relevant entities $e_i$ and relationships $r_j$ from situation triples:
    {\small
    \[
    E = \{e_i \mid (situation, has\_target\_entity, e_i)\}, \quad
    R = \{r_j \mid (situation, has\_intent, r_j)\}
    \]
    }
    \item Compute semantic similarity between extracted triples and \emph{Chronicle} embeddings to identify relevant nodes and relations:
     {\small
     \[
    sim(node_k) = semantic\_similarity(\text{embedding}(triples), \text{embedding}(node_k)), \forall node_k \in Chronicle
    \]
    }
    \item Formulate a query based on semantically aligned nodes and relationships:
    {\small
    \begin{verbatim}
        MATCH (u:User)-[:OWNS]->(c:CreditCard)-[:HAS_SPENDING]->(s:Spending)
        WHERE u.id = "user_123"
        RETURN s.category, s.amount
    \end{verbatim}
    }
\end{enumerate}

To infer that \texttt{table\_in\_front} refers to a specific anchor point (e.g., \texttt{anchor\_12}), the \textit{Reasoner} applies spatial reasoning over a symbolic reference $sr$, the set of available anchor points $AP = \{ap_1, ap_2, ..., ap_n\}$, and the bounding box of the user $B_{\text{user}}$ (provided by the \emph{Spatial Monitor}), where each $ap_i$ has corresponding textual descriptions $desc(ap_i)$ and spatial metadata. This process is as follows:
\begin{enumerate}

    \item Compute semantic similarity between $sr$ and each $desc(ap_i)$: 
    {\small
    \[
    sim(sr, ap_i) = semantic\_similarity(\text{embedding}(sr), \text{embedding}(desc(ap_i))), \forall ap_i \in AP
    \]
    }
    \item Retain anchor points with similarity above a threshold $\theta$:
    {\small
    \[
    AP_{\text{sem}} = \{ ap_i \in AP \mid sim(sr, ap_i) \geq \theta \}
   \]
    }
    \item Apply geometric reasoning upon the 3D positions $pos(ap_i)$ to select only those anchor points in front of the user:
  {\small
  \[
  AP_{\text{front}} = \{ ap_i \in AP_{\text{sem}} \mid pos(ap_i) \in \text{Front}(B_{\text{user}}) \}
  \]
    }
    \item Select the most semantically similar anchor point:
  \small{
  \[
  anchor^* = \arg\max_{ap_i \in AP_{\text{front}}} sim(sr, ap_i) \quad\quad \text{(e.g., } anchor^* = \texttt{anchor\_12} \text{)}
\]
  }

\end{enumerate}


\subsubsection{Step 4: Media Generation by Object Synthesizer} \
The \textit{Object Synthesizer} transforms the \textit{Reasoner}'s spatial and media-based output into structured visual media. It generates pie chart data by computing proportional values for each spending category (e.g., dining, traveling, etc.):

\small {
\[
pie\_chart\_data = \{ (category_j, \frac{amount_j}{\sum_k amount_k}) \}
\]
}

\normalsize\par
\subsubsection{Step 5: XR Rendering} \
This structured output is passed to the second LLM within the \textit{LLM Scribe Module}, which translates the pie chart configuration into an XR-executable script:
\vspace{-0.15cm}
{\small
\begin{verbatim}
{   event: "instantiate_visualization",
    visualization_type: "pie_chart",
    data: [{"category": "Dining", "amount": 320},
           {"category": "Travel", "amount": 210},
           {"category": "Groceries", "amount": 400}],
    position: anchor_12,
    interaction: enabled  }
\end{verbatim}
}
The \textit{XR Scene Generator} executes this script, rendering the pie chart on the designated anchor point in the 3D scene.

\subsubsection{Step 6: User Feedback Loop via User Updates} \
As the user interacts with the pie chart, the \textit{User State Monitoring} module captures behavioral cues (e.g., extended gaze, emotional responses), that are sent to the \textit{Reasoner}, refining user preferences in the \emph{Chronicle} and improving the personalization of future XR experiences.

{\small
\begin{verbatim}
<user, has_attention_on, pie_chart_001>
<user, has_emotion, curious>
\end{verbatim}
}


\subsection{Perspective-Aware Desk Environment}

This scenario demonstrates how PAiR enables OpenDome to provide hyper-personalized immersive desk experiences based on emotional cues and contextual reasoning, following the formalism from Section \ref{pafh}.



The interaction begins passively with the \textit{User State Monitor}, which gathers data for the \textit{Situation and Context Detectors}.
These detectors capture facial cues, for example, indicating sadness or longing:

\vspace{-0.2cm}
{\small
\begin{verbatim}
<user, has_emotion, sad>
<user, facial_expression, low_brows>
<user, gaze_direction, downward>
\end{verbatim}
}
\vspace{-0.2cm}

\begin{figure}[t]
  \centering
  \begin{minipage}[t]{0.49\textwidth}
    \centering
    \includegraphics[width=\textwidth]{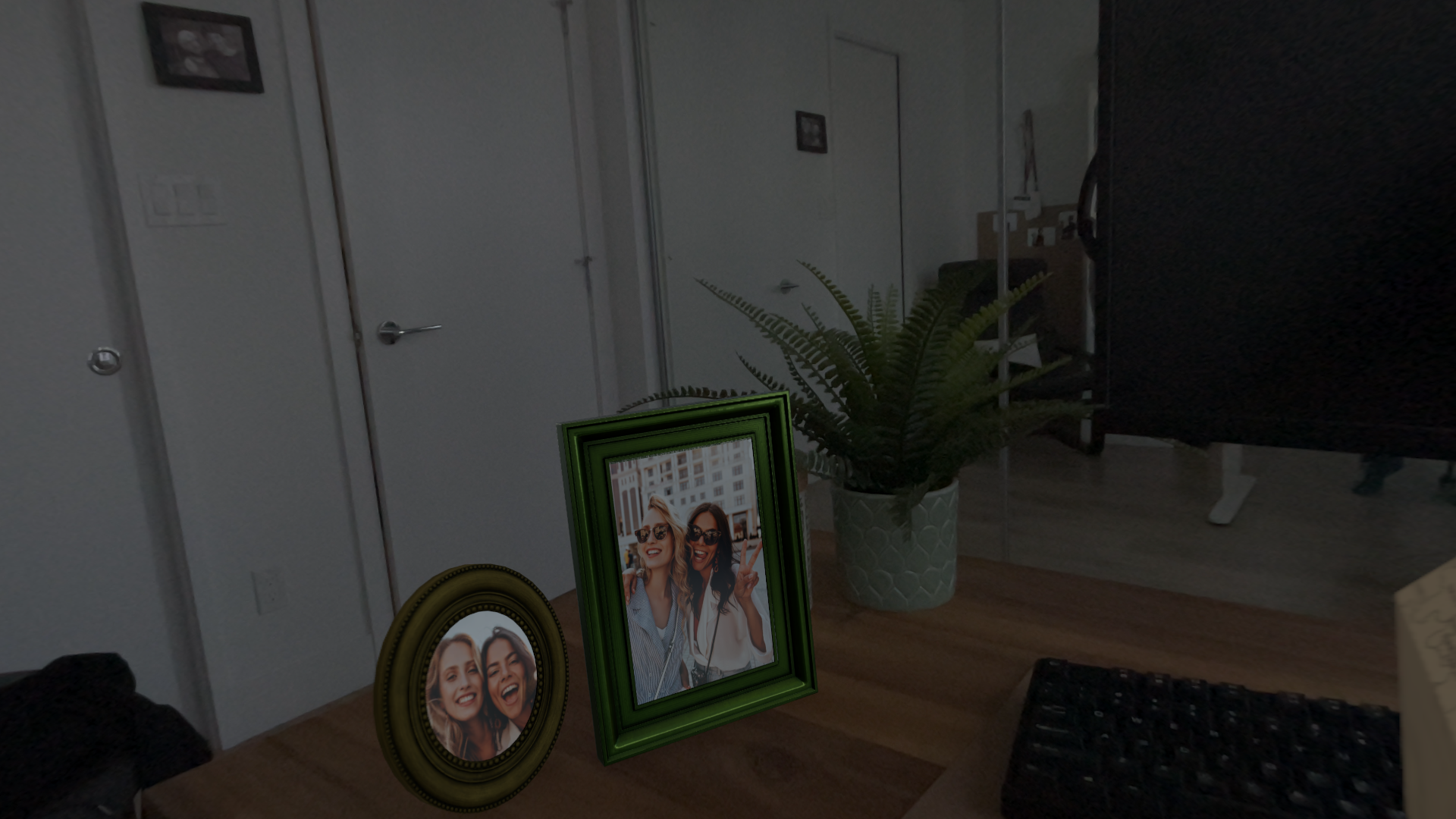}
    \caption{A perspective-aware desk environment powered by \emph{Chronicle}-driven semantic queries.}
    \label{fig:opendome2}
  \end{minipage}
  \hfill
  \begin{minipage}[t]{0.49\textwidth}
    \centering
    \includegraphics[width=\textwidth]{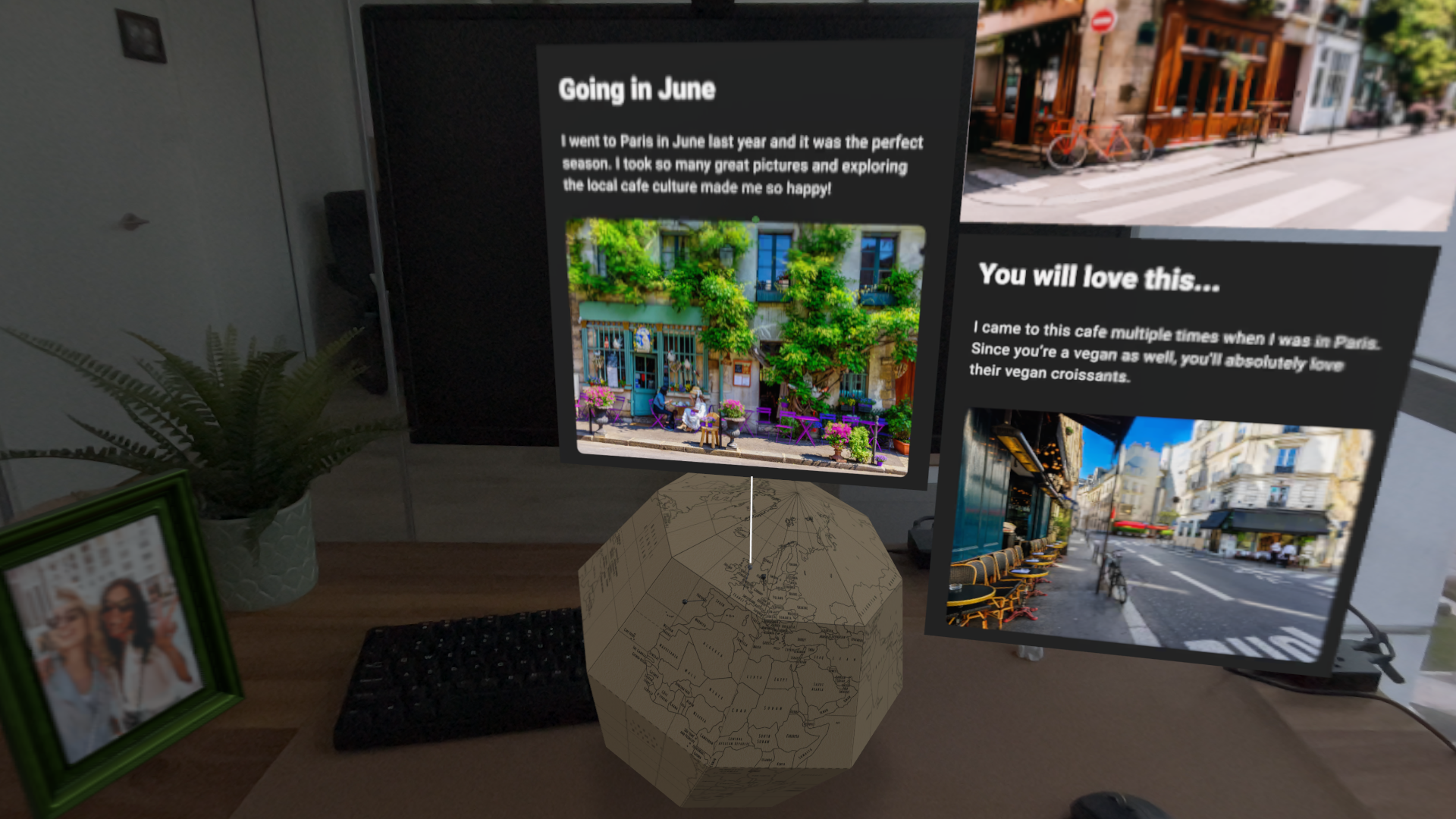}
    \caption{Example interaction between a user and another individual's \emph{Chronicle}.}
    \label{fig:opendome1}
  \end{minipage}
\end{figure}

These cues are interpreted semantically by the Reasoner LLM, resulting in:


{\small
\begin{verbatim}
<situation_2, has_participant, user>
<situation_2, has_emotion, sad>
<situation_2, has_possible_cause, missing_someone>
\end{verbatim}
}


The \textit{Reasoner} then identifies an emotional reference based on the user's \emph{Chronicle} (the user's past trip with their best friend) in attempt to increase the user's happiness (or influence another emotion, depending on the goal of the XR application). Next, the \emph{Reasoner} generates a semantic query to retrieve relevant visual data:

{\small
\begin{verbatim}
MATCH (u:User)-[:HAS_MEMORY]->(m:Memory)
WHERE m.context = "trip_with_best_friend"
RETURN m.image, m.location, m.sentiment
\end{verbatim}
}


The \textit{Reasoner} is aware of the environment's spatial setup as provided by the \textit{Spatial Monitor}, including anchor points and their descriptions:

{\small
\begin{verbatim}
<anchor_07, has_description, "frame holder for emotional memories">
<environment, supports, frame_color_customization>
\end{verbatim}
}



The \textit{Object Synthesizer} then jointly leverages spatial information and the user's preferences from their \emph{Chronicle} (e.g., favorite color: light blue) to prepare a personalized frame configuration. No assumption is made regarding the presence of a virtual picture frame; instead, ontological reasoning is applied to identify suitable anchor points based on available placements provided during the initialization phase.



The second fine-tuned LLM translates the symbolic output from the \textit{Reasoner} into XR-executable code:

{\small
\begin{verbatim}
{ "event": "instantiate_visualization",
  "visualization_type": "photo_frame",
  "data": [{ "image": "user_best_friend_berlin_trip.jpg", 
            "frame_color": "light_blue", 
            "emotional_tag": "nostalgia"}],
  "position": "anchor_07",
  "interaction": "enabled"  }
\end{verbatim}
}
The \textit{XR Scene Generator} then renders this output, placing the frame on the desk within the user's view (see Fig.~\ref{fig:opendome2}). As the user interacts with the personalized content, the \textit{User State Monitor} continues capturing feedback such as emotional uplift or focused attention, which can optionally be used to update their \emph{Chronicle}:


\vspace{-0.3cm}

\begin{verbatim}
<user, has_emotion, happy>
<user, has_attention_on, photo_frame_01>
\end{verbatim}

\section{Discussion \& Future Work}
\label{sec:discussion}

This paper introduced PAiR, a foundational framework that integrates Perspective-Aware AI with Extended Reality to enable dynamic, personalized, and shareable reasoning-ready immersive experiences. By grounding interactions in user-specific \emph{Chronicles} and leveraging symbolic reasoning, PAiR delivers semantically coherent and emotionally resonant XR content. Using OpenDome-based proof-of-concept scenarios, we showed that PAiR responds to both explicit prompts and passive emotional cues, enabling deeper human-centered intelligence.

Future work includes scaling \emph{Chronicles} to support multi-agent perspectives, enhancing various forms of reasoning, and integrating multimodal generative models for richer narrative synthesis. 
Furthermore, investigating real-time adaptation mechanisms in collaborative, multi-user XR environments—where participants’ perspectives may diverge—remains a critical direction for future research.







\begin{credits}
\subsubsection{\ackname} The authors would like to thank the team at Flybits, Toronto Metropolitan University, The Creative School, and MIT Media Lab for their support of our ongoing research in this field. 

\vspace{-0.2cm}
\subsubsection{\discintname}
The authors have no competing interests to declare that are relevant to the content of this article.
\end{credits}
%
%
%
 \bibliographystyle{splncs04}
 \bibliography{ref}

\begin{thebibliography}{10}
\providecommand{\url}[1]{\texttt{#1}}
\providecommand{\urlprefix}{URL }
\providecommand{\doi}[1]{https://doi.org/#1}

\bibitem{PAi1-2024}
Alirezaie, M., Rahnama, H., Pentland, A.: Structural learning in the design of perspective-aware ai systems using knowledge graphs. In: Proc. the AAAI- Digital Human Workshop (2024)

\bibitem{PAi-2021}
Rahnama, H., Alirezaie, M., Pentland, A.: A neural-symbolic approach for user mental modeling: A step towards building exchangeable identities. In: AAAI 2021 Symposium on Combining Machine Learning and Knowledge Engineering (2021)

\bibitem{PAi2-2024}
Alirezaie, M., Platnick, D., Rahnama, H., Pentland, A.: Perspective-aware ai (pai) for augmenting critical decision making. TechRxiv  (2024)

\bibitem{Sai2024twins}
Sai, S., Sharma, P., Gaur, A., Chamola, V.: Pivotal role of digital twins in the metaverse: A review. Digital Communications and Networks  (2024)

\bibitem{Radianti2020VR-Education}
Radianti, J., Majchrzak, T.A., Fromm, J., Wohlgenannt, I.: A systematic review of immersive virtual reality applications for higher education: Design elements, lessons learned, and research agenda. Computers \& Education  \textbf{147},  103778 (2020)

\bibitem{PAi3-2024}
Platnick, D., Alirezaie, M., Rahnama, H.: Enabling perspective-aware ai with contextual scene graph generation. Information  \textbf{15}(12), ~766 (2024)

\bibitem{AlhakamyXR-encompass}
Alhakamy, A.: Extended reality (xr) toward building immersive solutions: The key to unlocking industry 4.0. ACM Comput. Surv.  \textbf{56}(9) (Apr 2024)

\bibitem{Pasupuleti2025XRVRAR}
Pasupuleti, M.K.: Next-generation extended reality (xr): A unified framework for integrating ar, vr, and ai-driven immersive technologies. International Journal of Academic and Industrial Research Innovations  (03 2025)

\bibitem{Pat1999usermod}
Langley, P.: User modeling in adaptive interfaces. In: Proc. the Seventh Int. Conf. User Modeling. p. 357–370. UM '99, Springer-Verlag, Berlin, Heidelberg (1999)

\bibitem{Ratican2023-genai}
Ratican, J., Hutson, J., Wright, A.: A proposed meta-reality immersive development pipeline: Generative ai models and extended reality (xr) content for the metaverse. J. Intelligent Learning Systems and Applications  \textbf{15}(1),  24--35 (2023)

\bibitem{Schwarz2025meta}
Schwarz, K., Rozumnyi, D., Bulò, S.R., Porzi, L., Kontschieder, P.: A recipe for generating 3d worlds from a single image (2025)

\bibitem{10.3389/frvir.2022.1015155}
Germanakos, P., Sotirakou, C., Mourlas, C.I., Richir, S., Boomgaarden, H.: Editorial: Immersive reality and personalized user experiences. Frontiers in Virtual Reality  \textbf{3} (2022)

\bibitem{10.1007/2023}
Adhanom, I., MacNeilage, P., Folmer, E.: Eye tracking in virtual reality: a broad review of applications and challenges. Virtual Reality  \textbf{27},  1481–1505 (2023)

\bibitem{GAFFINET2025104230}
Gaffinet, B., {Al Haj Ali}, J., Naudet, Y., Panetto, H.: Human digital twins: A systematic literature review and concept disambiguation for industry 5.0. Computers in Industry  \textbf{166},  104230 (2025)

\bibitem{57dca900c67c4572bc5c185827d53c63}
Bozkir, E., Özdel, S., Lau, K.H.C., Wang, M., Gao, H., Kasneci, E.: Embedding large language models into extended reality: Opportunities and challenges for inclusion, engagement, and privacy. In: Proc. the 6th Conference on ACM Conversational User Interfaces, CUI 2024. Association for Computing Machinery (2024)

\bibitem{Jiang2024HAI}
Jiang, T., Sun, Z., Fu, S., Lv, Y.: Human-ai interaction research agenda: A user-centered perspective. Data and Information Management p. 100078 (2024)

\bibitem{Amershi2019Guidelines}
Amershi, S., Weld, D., Vorvoreanu, M., Fourney, A., Nushi, B., Collisson, P., Suh, J., Iqbal, S., Bennett, P.N., Inkpen, K., Teevan, J., Kikin-Gil, R., Horvitz, E.: Guidelines for human-ai interaction. In: Proc. the 2019 CHI Conf. Human Factors in Computing Systems. pp. 1--13. ACM, New York, NY, USA (2019)

\bibitem{vanDerMeulen2025ToM}
van~der Meulen, R., Verbrugge, R., van Duijn, M.: Towards properly implementing theory of mind in ai: An account of four misconceptions (2025), arXiv:2503.16468

\bibitem{Kosinski2013DigitalFootprint}
Kosinski, M., Stillwell, D., Graepel, T.: Private traits and attributes are predictable from digital records of human behavior. Proc. the National Academy of Sciences  \textbf{110}(15),  5802--5805 (2013)

\bibitem{raffel2023exploringlimitstransferlearning}
Raffel, C., Shazeer, N., Roberts, A., Lee, K., Narang, S., Matena, M., Zhou, Y., Li, W., Liu, P.J.: Exploring the limits of transfer learning with a unified text-to-text transformer (2023), \url{https://arxiv.org/abs/1910.10683}

\bibitem{Graessler2017DigitalTwinHuman}
Graessler, I., Pohler, A.: Integration of a digital twin as human representation in a scheduling procedure of a cyber-physical production system. In: Proceedings of the 2017 IEEE International Conference on Industrial Engineering and Engineering Management (IEEM). pp. 289--293. Singapore (2017)

\bibitem{Marougkas2024VRPersonalization}
Marougkas, A., Troussas, C., Krouska, A., Sgouropoulou, C.: How personalized and effective is immersive virtual reality in education? a systematic literature review for the last decade. Multimedia Tools and Applications  \textbf{83},  18185--18233 (2024)

\bibitem{Pardini2022PersonalizedVR}
Pardini, S., Gabrielli, S., Dianti, M., Novara, C., Zucco, G.M., Mich, O., Forti, S.: The role of personalization in the user experience, preferences and engagement with virtual reality environments for relaxation. International Journal of Environmental Research and Public Health  \textbf{19}(12), ~7237 (2022)

\bibitem{vanLoon2018VR-Empathy}
Van~Loon, A., Bailenson, J., Zaki, J., Bostick, J., Willer, R.: Virtual reality perspective-taking increases cognitive empathy for specific others. PLoS ONE  \textbf{13}(8) (2018)

\bibitem{Tehranchi2023CognitiveUI}
Tehranchi, F., Ritter, F.E.: A user model to directly compare two unmodified interfaces: a study of including errors and error corrections in a cognitive user model. AI EDAM (Artificial Intelligence for Engineering Design, Analysis and Manufacturing)  \textbf{37}(2),  248--265 (2023)

\end{thebibliography}

\end{document}